\documentclass[runningheads]{llncs}
\usepackage{graphicx}

\usepackage{tikz}
\usepackage{comment}
\usepackage{amsmath,amssymb} 
\usepackage{color}
\usepackage{orcidlink}
\usepackage{floatrow}
\usepackage{multirow}
\usepackage{threeparttable}

\usepackage[accsupp]{axessibility}  

\usepackage{xspace}
\def\onedot{.\xspace}
\def\eg{\emph{e.g}\onedot} 

\def\ie{\emph{i.e}\onedot}

\def\etc{\emph{etc}\onedot}

\newfloatcommand{figurebox}{figure}[\nocapbeside][\dimexpr(\textwidth-\columnsep)/2\relax]
\newfloatcommand{tablebox}{table}[\nocapbeside][\dimexpr(\textwidth-\columnsep)/2\relax]

\begin{document}
\pagestyle{headings}
\mainmatter
\def\ECCVSubNumber{6686}  

\title{VSA: Learning Varied-Size Window Attention in Vision Transformers} 

\titlerunning{VSA}
%
\author{
Qiming Zhang\inst{1}\orcidlink{0000-0003-0060-0543} \thanks{Equal contribution.} \and
Yufei Xu\inst{1}\orcidlink{0000-0002-9931-5138}* \and
Jing Zhang\inst{1}\orcidlink{0000-0001-6595-7661} \and 
Dacheng Tao\inst{2,1}\orcidlink{0000-0001-7225-5449}}

\authorrunning{Q. Zhang et al.}
%
\institute{University of Sydney, Australia \and
JD Explore Academy, China \\
\email{\{yuxu7116,qzha2506\}@uni.sydney.edu.au, \\ jing.zhang1@sydney.edu.au, dacheng.tao@gmail.com}}

\maketitle

\begin{abstract}
Attention within windows has been widely explored in vision transformers to balance the performance, computation complexity, and memory footprint. However, current models adopt a hand-crafted fixed-size window design, which restricts their capacity of modeling long-term dependencies and adapting to objects of different sizes. To address this drawback, we propose \textbf{V}aried-\textbf{S}ize Window \textbf{A}ttention (VSA) to learn adaptive window configurations from data. Specifically, based on the tokens within each default window, VSA employs a window regression module to predict the size and location of the target window, \ie, the attention area where the key and value tokens are sampled. By adopting VSA independently for each attention head, it can model long-term dependencies, capture rich context from diverse windows, and promote information exchange among overlapped windows. VSA is an easy-to-implement module that can replace the window attention in state-of-the-art representative models with minor modifications and negligible extra computational cost while improving their performance by a large margin, e.g., 1.1\% for Swin-T on ImageNet classification. In addition, the performance gain increases when using larger images for training and test. Experimental results on more downstream tasks, including object detection, instance segmentation, and semantic segmentation, further demonstrate the superiority of VSA over the vanilla window attention in dealing with objects of different sizes. The code is available at \href{https://github.com/ViTAE-Transformer/ViTAE-VSA}{https://github.com/ViTAE-Transformer/ViTAE-VSA}.
\end{abstract}

\section{Introduction}
Recent Vision transformers have shown great potential in various vision tasks. By stacking multiple transformer blocks with vanilla attention, ViT~\cite{vit} processes non-overlapping image patches and obtain superior classification performance. However, vanilla attention with quadratic complexity over the input length is hard to adapt to vision tasks with high-resolution images as input due to the expensive computational cost. To alleviate such issues, window-based attention~\cite{liu2021swin} is proposed to partition the images into local windows and conduct attention within each window to balance the performance, computation complexity, as well as memory footprint. This mechanism enables vision transformers to make a great success in many downstream visual tasks~\cite{liu2021swin,yang2021focal,dong2021cswin,zhang2022vitaev2,yang2022modeling,xu2021vitae,xu2022vitpose,wang2021kvt}. However, it also enforces a spatial constraint on transformers' attention distance, i.e., within the predefined window at each layer, thereby limiting the transformer's ability to deal with objects at different scales.

Recent works have explored heuristic designs of attending to more tokens to alleviate such a spatial constraint. For example, Swin transformer~\cite{liu2021swin} enlarges the window sizes from 7 $\times$ 7 to 12 $\times$ 12 when varying the image size from 224 $\times$ 224 to 384 $\times$ 384, and sets the window size as 32 $\times$ 32 to deal with image size 640 $\times$ 640 in SwinV2~\cite{swinv2}. Some other methods try to find a good trade-off between attending to more tokens and increasing attention distance, \eg, multiple window mechanisms have been explored in Focal attention \cite{yang2021focal}, where coarse granularity tokens are involved in capturing long-distance information. Cross-shaped window attention~\cite{dong2021cswin} relaxes the spatial constraint of the window in vertical and horizontal directions and allows the transformer to attend to far-away relevant tokens along with the two directions while keeping the constraint along the diagonal direction. Pale~\cite{wu2021pale} further increases the diagonal-direction attention distance by attending to tokens in the dilated vertical/horizontal directions. These methods have achieved superior performance in image classification tasks by enlarging the attention distance. However, they sacrifice computational efficiency and consume more memory, especially when training large models with high-resolution images. Besides, all these methods determine the window sizes heuristically. Intuitively, using a fixed-size window may be sub-optimal for dealing with objects of different sizes, although stacking more layers could mitigate this issue to some extent, which may also result in more parameters and optimization difficulty. In this paper, we argue that if the window can be relaxed to a varied-size rectangular one, whose size and position are learned directly from data, the transformer can capture rich context from diverse windows and learn more powerful object feature representation.

\thisfloatsetup{heightadjust=all,valign=c}
\begin{figure}[t!]
    \begin{floatrow}[2]
      \ffigbox[1.5\FBwidth]
      {\includegraphics[width=1.\linewidth]{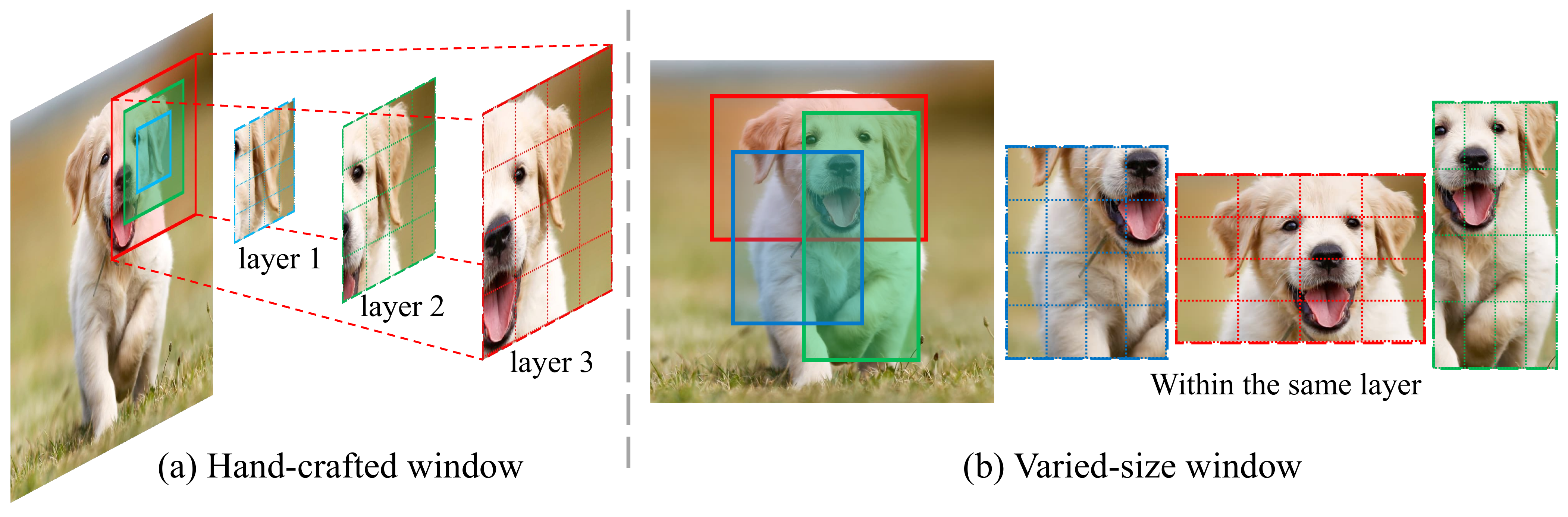}}
      {\caption{The comparison of the current works (hand-crafted windows) and the proposed VSA (varied-size windows).}\label{fig:beginning}}%
      \ffigbox[0.5\FBwidth]
      {\caption{The performance with different image sizes.}\label{fig:performance}}
      {\includegraphics[width=\linewidth]{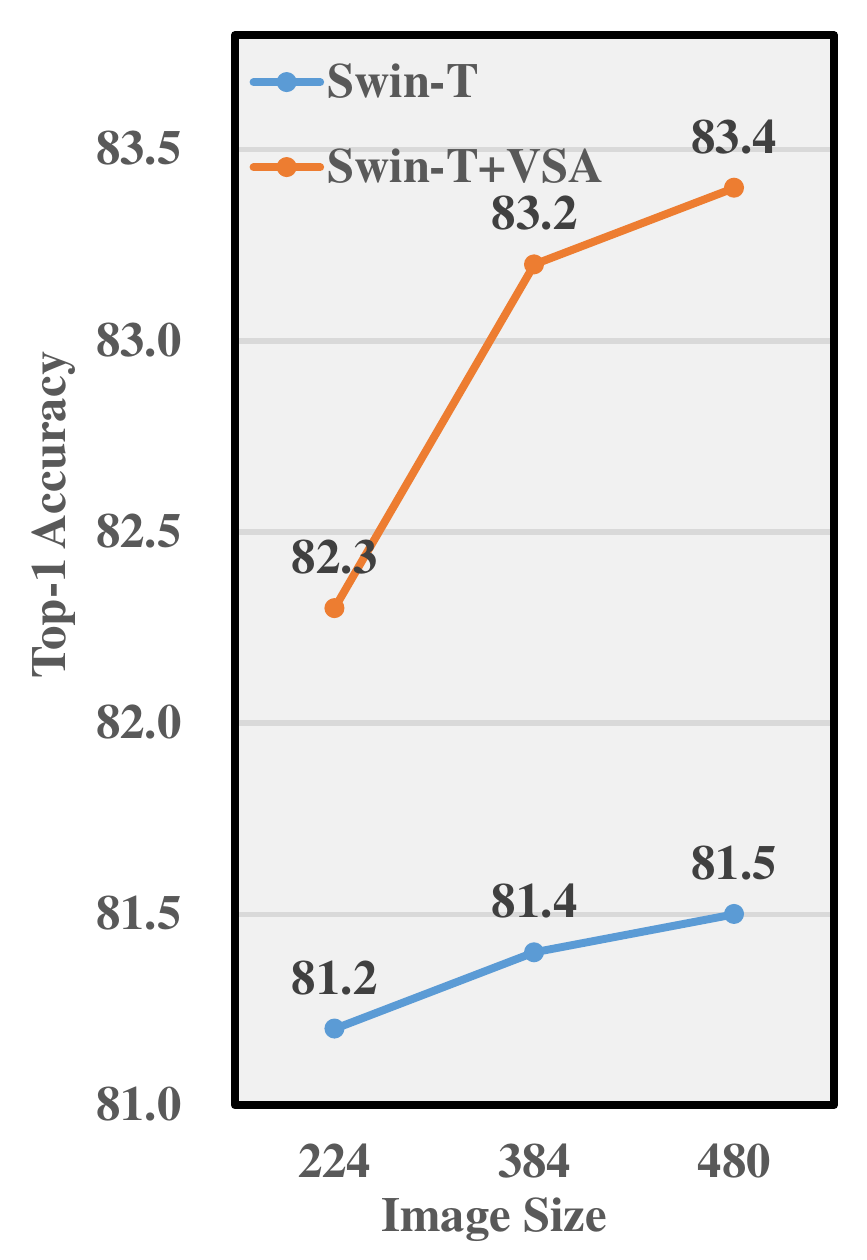}}
    \end{floatrow}
\end{figure}

To this end, we propose a novel \textbf{V}aried-\textbf{S}ize Window \textbf{A}ttention (VSA) mechanism to learn adaptive window configurations from data. Different from the previous window-based transformers where query, key, and value tokens are all sampled from the same window as shown in Figure~\ref{fig:beginning}(a), VSA employs a window regression module to predict the size and location of the target window based on the tokens within each default window. Then, the key and values tokens are sampled from the target window. By adopting VSA independently for each attention head, it enables the attention layers to model long-term dependencies, capture rich context from diverse windows, and promote information exchange among overlapped windows, as illustrated in Figure~\ref{fig:beginning}(b). VSA is an easy-to-implementation module that can replace the window attention in state-of-the-art representative models with minor modifications and negligible extra computational cost while improving their performance by a large margin, e.g., 1.1\% for Swin-T on ImageNet classification. In addition, the performance gain increases when using larger images for training and test, as shown in Figure~\ref{fig:performance}. With the larger images as input, Swin-T with predefined window sizes cannot adapt to large objects well, and the improvement brought by enlarging image sizes is marginal, i.e., a gain of 0.3\% from 224 $\times$ 224 to 480 $\times$ 480. In contrast, the performance gain of VSA over Swin-T increases significantly from 1.1\% to 1.9\%, owing to the varied-size window attention. Besides, as VSA can effectively promote information exchange across overlapped windows via token sampling, it does not need the shifted windows mechanism in Swin.

In conclusion, the contribution of this study is threefold. (1) We introduce a novel VSA mechanism that can directly learn adaptive window size and location from data. It breaks the spatial constraint of the fixed-size window in existing works and makes it easier for window-based transformers to adapt to objects at different scales. (2) VSA can serve as an easy-to-implement module to improve various window-based transformers, including but not limited to Swin~\cite{liu2021swin,swinv2} and ViTAEv2~\cite{xu2021vitae,zhang2022vitaev2}, with minor modifications and negligible extra computational cost. (3) Extensive experimental results on public benchmarks demonstrate the superiority of VSA over the vanilla window attention on various visual tasks, including image classification, object detection, and semantic segmentation.

\section{Related Work}

\subsection{Window-based vision transformers}

Vision transformers~\cite{vit} have demonstrated superior performance in many vision tasks by modeling long-term dependencies among local image patches (a.k.a. tokens)~\cite{xu2022vitpose,jing2020dynamic}. However, vanilla full attention performs poorly in training efficiency due to the shortage of inductive bias. To improve the efficiency, the following works either implicitly or explicitly introduce inductive bias into vision transformers~\cite{touvron2020training,xu2021vitae,dai2021coatnet,yan2021contnet} and obtain superior classification performance. After that, multi-stage design has been explored in \cite{wang2021pyramid,wang2021pvtv2,liu2021swin,wang2021crossformer,zhang2022vitaev2} to better adapt vision transformers to downstream vision tasks. Among them, Swin~\cite{liu2021swin} is a representative work. By partitioning the tokens into non-overlapping windows and conducting attention within each window, Swin alleviates the huge computational cost caused by attention when dealing with larger input images. Although it balances the performance, computational cost, and memory footprints well, window-based attentions bring a spatial constraint on the attention distance due to the constant maximum size of windows. To alleviate such issues, different techniques have been explored to recover the transformer's ability to model long-term dependency gradually, \eg, using additional tokens for efficient cross-window feature exchange or designing delicate windows to allow the transformer layers to attend to far-away tokens in specific directions~\cite{fang2021msg,dong2021cswin,wu2021pale,huang2021shuffle}. However, they still 1) rely on heuristic-designed windows for attention computation and 2) need to stack the transformers layers sequentially to enable feature exchange across all windows and model long-term dependencies. Thus, they lack the flexibility to adapt well to inputs of various sizes since their maximum attention distances are restricted by the constant and data-agnostic window size and model depth.

Unlike them, the proposed VSA estimates window sizes and locations adaptively based on input features and calculates attention within such windows. Therefore, VSA allows transformer layers to model long-term dependencies, capture rich context, and promote cross-window information exchange from diverse varied-size windows. As VSA learns the window sizes in a data-driven manner, it can benefit window-based vision transformers to adapt to objects at various scales and thus helps boost their performance on image classification, object detection, and semantic segmentation.

\subsection{Deformable sampling}

Deformable sampling has been widely explored previously to help the convolution networks~\cite{dai2017deformable,zhu2019deformable} to focus on regions of interest and extract better features. Similar mechanisms have been exploited in deformable-DETR~\cite{zhu2021deformable} to help the transformer detector to find and utilize the most valuable token features for object detection in a sparse manner. Recently, DPT~\cite{chen2021dpt} designs deformable patch merging layers based on PVT~\cite{wang2021pyramid} to help the transformer to preserve better features after downsampling. VSA, from another perceptive, introduces learnable varied-size window attention into transformers. By flexibly estimating the window sizes and locations for attention calculation, VSA breaks the spatial constraint of fixed-size windows and makes it easier for window-based transformers to better adapt to the objects at various scales. 

\section{Method}

In this section, we will take Swin transformer~\cite{liu2021swin} as an example and give a detailed description of applying VSA in Swin. The details of incorporating VSA into ViTAE~\cite{zhang2022vitaev2} will be presented in the supplementary.

\subsection{Preliminary}
We will first briefly review the window attention operation in the baseline method Swin transformer. Given the input features $X \in \mathcal{R}^{H \times W \times C}$ as input, Swin transformer employs several window-based attention layers for feature extraction. In each window-based attention layer, the input features are firstly partitioned into several non-overlapping windows, \ie, $\{X_w^i \in \mathcal{R}^{w \times w \times C} | i \in [1, \dots, \frac{H\times W}{w^2}]\}$, where $w$ is the predefined window size. After that, the partitioned tokens are flatten along the spatial dimension and projected to query, key, and value tokens, \ie, $\{Q_{w,f}^i, K_{w,f}^i, V_{w,f}^i \in \mathcal{R}^{w^2 \times N \times C'} | i \in [1, \dots, \frac{H\times W}{w^2}]\}$, where $Q,K,V$ represent the query, key, and value tokens, respectively, $N$ denotes the head number and $C'$ is the channel dimension along each head. It is noted that $N \times C'$ equals the channel dimension $C$ of the given feature. Given the flattened query, key, and value tokens from the same default window, the window-based attention layers conduct full attention within the window, \ie,
\begin{equation}
    F_{w,f}^i = MHSA(Q_{w,f}^i, K_{w,f}^i, V_{w,f}^i).
\end{equation}
The $F_{w,f}^i \in \mathcal{R}^{w^2\times N \times C'}$ is the features after attention and $MHSA$ represents the vanilla multi-head self-attention operation~\cite{vit}. The relative position embeddings are utilized during the attention calculation to encode spatial information into the features. The extracted features $F$ are reshaped back to the window shape, \ie, $F_w^i \in \mathcal{R}^{w \times w \times C}$, and added with the input feature $X_w^i$. The same operation is individually repeated for each window and the generated features from all windows are then concatenated to recover the shape of input features. After that, an FFN module is employed to refine the extracted features, which contains two linear layers with hidden dimension $\alpha C$, where $\alpha$ is the expansion ratio. For notation simplification, we dismiss the window index notation $i$ in the following since each window's operations are the same.

With the usage of window-based attention, the computational complexity decreases to linear to the input size, \ie, each window attention's complexity is $\mathcal{O}(w^4C)$ and the computation complexity of window attention for each image is $\mathcal{O}(w^2HWC)$. To bridge connections between different windows, shifted operations are used between two adjacent transformer layers in Swin~\cite{liu2021swin}. As a result, the receptive field of the model is gradually enlarged with layers stacking in sequence. However, current window-based attentions restrict the attention area of the tokens within the corresponding hand-crafted window at each transformer layer. It limits the model's ability to capture far-away contextual information and learn better feature representations for objects at different scales.

\begin{figure}
    \centering
    \includegraphics[width=\linewidth]{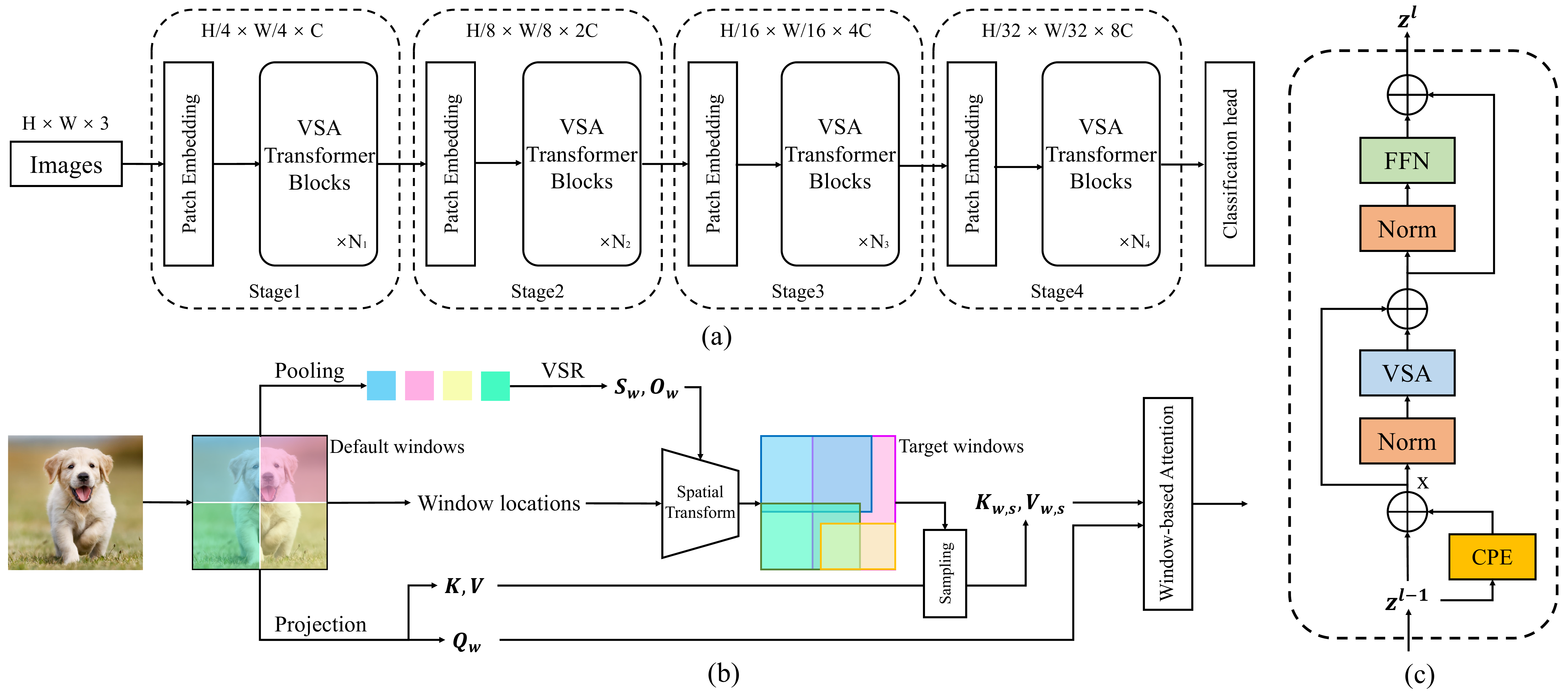}
    \caption{The pipeline of the transformer with our proposed varied-size window attention. (a) The overall structure of stacking VSA transformers blocks; (b) The details of the proposed VSA module; (c) The pipeline of the VSA transformer block.}
    \label{fig:pipeline}
\end{figure}

\subsection{Varied-size window attention}
\noindent\textbf{Base window generation.}
Rather than stacking layers with hand-crafted windows to gradually enlarge the receptive field, our VSA allows the query tokens to attend to far-away regions and empower the network with the flexibility to determine the target window size, \ie, attention area, given specific input data at each layer. VSA only needs to make minor modifications to the basic structure of backbone networks and serves as an easy-to-implement module to replace the vanilla window attention in window-based transformers as in Figure~\ref{fig:pipeline} (a). Technically, given the input features $X$, VSA first partitions these tokens into several windows $X_w$ with the predefined window size $w$, following the baseline methods' routine. We refer to these windows as default windows and get the query features from the default windows, \ie, 
\begin{equation}
    Q_w = Linear(X_w).
\end{equation}
\noindent\textbf{Varied-size window regression module.}
To estimate the size and location of the target window for each default window, VSA considers the size and location of the default window as a reference and adopts a varied-size window regression ($VSR$) module to predict the scale and offset upon the references as shown in Figure~\ref{fig:pipeline}(b). The $VSR$ module consists of an average pooling layer, a LeakyReLU~\cite{xu2015empirical} activation layer, and a $1 \times 1$ convolutional layer with stride 1 in sequence. The kernel size and stride of the pooling layer follow the default window size, \ie,
\begin{equation}
    S_w, O_w = Conv \circ LeakyReLU \circ AveragePool(X_w),
\end{equation}
where $S_w$ and $O_w \in \mathcal{R}^{2 \times N}$ represent the estimated scales and offsets in the horizontal and vertical directions w.r.t. the default window locations, independently for $N$ attention heads. The generated windows are referred to as target windows.

\noindent\textbf{Varied-size window-based attention.}
We first get the key and value tokens $K,V \in \mathcal{R}^{H\times W \times C}$ from the feature map $X$, \ie, 
\begin{equation}
    K, V = Reshape \circ Linear(X).
\end{equation}
Then the VSA module uniformly samples $M$ features from each varied-size window over $K,V$ respectively, and obtains $K_{w,v},V_{w,v} \in \mathcal{R}^{ M \times N \times C'}$ to serve as the key/value tokens for the query tokens $Q_w$. To keep the computational cost as window attention, we set $M$ equal to $w \times w$. The sampled tokens $K_{w,v},V_{w,v}$ are then fed into $MHSA$ with queries $Q_{w}$ for attention calculation. However, as the key/value tokens are sampled from different locations with the query tokens, the relative position embeddings between the query and key tokens may not describe the spatial relationship well. Following the spirit in CPVT~\cite{chu2021conditional}, we adopt conditional position embedding (CPE) before the MHSA layers to supply the spatial relationships into the model as shown in Figure~\ref{fig:pipeline} (c), \ie, 
\begin{equation}
    X = Z^{l-1} + CPE(Z^{l-1}),
\end{equation}
where $Z^{l-1}$ is the feature from the previous transformer block and $CPE$ is implemented by a depth-wise convolution layer with kernel size equal the window size, \ie, $7 \times 7$ by default, and stride 1.

\subsection{Computation complexity analysis}
The extra computations caused by VSA come from the $CPE$ and $VSR$ module, while the other parts, including the window-based multi-head self-attention and FFN network, are exactly the same as the baseline models. Given the input features $X \in \mathcal{R}^{H \times W \times C}$, VSA firstly uses a depth-wise convolutional layer with $7 \times 7$ kernels to generate CPE, which brings extra $\mathcal{O}(49\cdot HWC)$ computations. In the $VSR$ module, we first employ an average pooling layer with kernel size and stride equal to the window size to aggregate features from the default windows, whose complexity is $\mathcal{O}(HWC)$. 
The following activation function does not introduce extra computations, and the last convolutional layer with kernel size $1 \times 1$ takes $X_{pool} \in \mathcal{R}^{\frac{H}{w} \times \frac{W}{w} \times C}$ as the input and estimates the scales $S_w$ and offsets $O_w$. Both the scales and offsets belong to $\mathcal{R}^{2 \times N}$. Thus, the computational complexity of the convolutional layer is $\mathcal{O}(\frac{4N}{w^2} H W C)$,
where $N$ is the number of the attention heads in the transformer layers, and $w$ is the window size. After obtaining the scales and offsets, we transform the default windows to the varied-size windows and uniformly sample $w \times w$ tokens within each target window. The computational complexity for each window is $w^2 \times 4 \times C$, and the total computational complexity for the sampling operation is $\mathcal{O}(4\cdot H W C)$.
Thus, the total extra computations brought by VSA is $\mathcal{O}\{(54 + \frac{4N}{w^2}) H W C\}$, which is far less ($\leq5\%$) than the total computational cost of the baseline models, regarding the complexity of FFN is $\mathcal{O}(2 \alpha H W C^2)$ and $C$ is always larger than 96.

\section{Experiments}

\subsection{Implementation details}
We evaluate the performance of the proposed VSA based on Swin~\cite{liu2021swin} and ViTAEv2~\cite{zhang2022vitaev2}. The former is a pure transformer model with shifted windows between two adjacent layers, while the latter is an improved transformer model by introducing convolution inductive bias, which models long- and short-term dependencies jointly. In this paper, we adopt the full-window version of ViTAEv2 as the baseline. All the models are trained for 300 epochs from scratch on the standard ImageNet-1k~\cite{deng2009imagenet} dataset with input resolution $224 \times 224$. We follow the hyper-parameters setting in the baseline methods to train the variants with VSA, \eg, we use AdamW~\cite{loshchilov2018decoupled} optimizer with cosine learning rate schedulers during training. A 20-epochs linear warm-up is utilized following Swin~\cite{liu2021swin} to stabilize training. The initial learning rate is set to 0.001 for 1024 batch size during training. The data augmentation is the same as \cite{liu2021swin} and \cite{zhang2022vitaev2}, \ie, random cropping, auto-augmentation~\cite{autoaug}, CutMix~\cite{yun2019cutmix}, MixUp~\cite{zhang2017mixup}, and random erasing are used to augment the input images. Besides, label smoothing with a weight of 0.1 is adopted. It is also noteworthy that there is no shifted window mechanism in the models with VSA, since VSA enables cross-window information exchange among overlapped varied-size windows.

\subsection{Image Classification on ImageNet}

We evaluate the classification performance of different models on the ImageNet~\cite{deng2009imagenet} validation set. As shown in Table~\ref{tab:Classification}, the proposed VSA helps boost the classification accuracy of Swin transformer by 1.1\% absolute Top-1 accuracy, \ie from 81.2\% to 82.3\%, even without the shifted window mechanisms. It indicates that VSA can flexibly determine the appropriate window sizes and locations given the input features, allow the tokens to effectively attend far-away but relevant tokens outside the default windows to extract rich context, and learn better feature representations.
Besides, Swin-T with VSA obtains comparable performance with MSG-T~\cite{fang2021msg}, which adopts extra messenger tokens for feature exchange across windows, \ie, 82.3\% v.s. 82.4\%, demonstrating that our varied-size window mechanism can enable sufficient feature exchange across windows without the need of using extra tokens. For ViTAEv2~\cite{zhang2022vitaev2}, ViTAEv2-S with VSA obtains 82.7\% (+0.5\%) classification accuracy with only 20M parameters, demonstrating that the proposed varied-size window attention is compatible with not only the transformers with vanilla window attentions but also those with convolutions for feature exchange across windows. 

\begin{table}[h]
  \centering
  \caption{Image classification results on ImageNet. `Input Size' denotes the image size used for training and test.}
  \scriptsize
  \begin{threeparttable}
    \begin{tabular}{l|ccc|cc|c}
    \hline
    \multicolumn{1}{c|}{\multirow{2}[2]{*}{Model}} & Params & FLOPs & Input & \multicolumn{2}{c|}{ImageNet~\cite{deng2009imagenet}} & Real~\cite{beyer2020we} \\
          & (M)   & (G)   & Size  & Top-1 & Top-5 & Top-1 \\
    \hline
    DeiT-S~\cite{touvron2020training} & 22  & 4.6   & 224   & 81.2  & 95.4  & 86.8  \\
    PVT-S~\cite{wang2021pyramid} & 25  & 3.8   & 224   & 79.8  & - & -  \\
    ViL-S~\cite{zhang2021multi} & 25 & 4.9 & 224 & 82.4 & - & - \\
    PiT-S~\cite{heo2021pit} & 24  & 4.8   & 224   & 80.9  &   -    & - \\
    TNT-S~\cite{han2021transformer} & 24  & 5.2   & 224   & 81.3  & 95.6  & - \\
    MSG-T~\cite{fang2021msg} & 25 & 3.8 & 224 & 82.4 & - & - \\
    Twins-PCPVT-S~\cite{chu2021twins} & 24  & 3.8   & 224   & 81.2  &    -   &  - \\
    Twins-SVT-S~\cite{chu2021twins} & 24  & 2.9   & 224   & 81.7  & -      & - \\
    T2T-ViT-14~\cite{yuan2021tokens} & 22  & 5.2   & 224   & 81.5  & 95.7  & 86.8  \\
    Swin-T~\cite{liu2021swin} & 29  & 4.5   & 224   & 81.2  &    -   & - \\
    \textbf{Swin-T$+$VSA} & \textbf{29} & \textbf{4.6} & \textbf{224} & \textbf{82.3} & \textbf{96.1}  & \textbf{87.5} \\
    ViTAEv2-S\footnotemark[1]~\cite{zhang2022vitaev2} &  20  & 5.4  & 224   & 82.2  & 96.1  & 87.5 \\
    \textbf{ViTAEv2-S\footnotemark[1]$+$VSA} &  \textbf{20}   &  \textbf{5.6} & \textbf{224} & \textbf{82.7} &  \textbf{96.3} & \textbf{87.8} \\
    \hline
    Swin-T~\cite{liu2021swin} & 29  &  14.2  & 384   & 81.4  &  95.4 & 86.4 \\
    \textbf{Swin-T$+$VSA} & \textbf{29} & \textbf{14.9} & \textbf{384} & \textbf{83.2} & \textbf{96.5}  & \textbf{88.0} \\
    Swin-T~\cite{liu2021swin} & 29  &  23.2  & 480   & 81.5  &  95.7 & 86.3 \\
    \textbf{Swin-T$+$VSA} & \textbf{29} & \textbf{24.0} & \textbf{480} & \textbf{83.4} & \textbf{96.7} & \textbf{88.0}  \\
    \hline
    PiT-B~\cite{heo2021pit} & 74  & 12.5  & 224   & 82.0  &   -    & - \\
    TNT-B~\cite{han2021transformer} & 66  & 14.1  & 224   & 82.8  & 96.3  & - \\
    Focal-B~\cite{yang2021focal} & 90  & 16.0  & 224   & 83.8  &  -     & - \\
    ViL-B~\cite{zhang2021multi} & 56 & 13.4 & 224 & 83.7 & - & - \\
    MSG-S~\cite{fang2021msg} & 56 & 8.4 & 224 & 83.4 & -  & - \\
    PVTv2-B5~\cite{wang2021pvtv2} & 82  & 11.8  & 224   & 83.8  &  -     & - \\
    Swin-S~\cite{liu2021swin} & 50  & 8.7   & 224   & 83.0  &   -    &  -\\
  \textbf{Swin-S$+$VSA} & \textbf{50} & \textbf{8.9} & \textbf{224} & \textbf{83.8} &  \textbf{96.8} & \textbf{88.54} \\
    \hline
    Swin-B~\cite{liu2021swin} & 88  & 15.4  & 224   & 83.3  &  - & 88.0 \\
    \textbf{Swin-B$+$VSA} & 88 & 16.0 & 224 & 83.9 & 96.7 & 88.6 \\
    \hline
    \end{tabular}%
  \label{tab:Classification}%
    \begin{tablenotes}
    \scriptsize
    \item[1] The full window version.
  \end{tablenotes}
  \end{threeparttable}
\end{table}%

When scaling the input images to higher resolutions, \ie, from 224 $\times$ 224 to 384 $\times$ 384 and 480 $\times$ 480, the performance gains from VSA become larger owing to its ability to learn adaptive target window sizes from data. Specifically, the performance gain brought by VSA increases from 1.1\% to 1.8\% absolute accuracy over Swin-T when scaling the input size from 224 to 384, respectively. For the 480 $\times$ 480 input resolution, the performance gain of VSA further increases to 1.9\%, while the Swin transformers only benefit from the higher resolution marginally (\ie, 0.2\%). The reason is that the fixed-size window attention in Swin limits the attention region at each transformer layer, which brings difficulty in handling objects at different scales. In contrast, VSA can learn to vary the window size to adapt to the objects and capture rich contextual information from different attention heads at each layer, which is beneficial for learning powerful object feature representations.

\subsection{Object detection and instance segmentation on MS COCO}

\textbf{Settings.} We evaluate the backbone models for the object detection and instance segmentation tasks on the MS COCO~\cite{lin2014microsoft} dataset, which contains 118K training, 5K validation, and 20K test images with full annotations. We adopt the models trained on ImageNet with 224 $\times$ 224 input resolutions as backbones and use three typical object detection frameworks, \ie, the two-stage frameworks Mask RCNN~\cite{he2017mask} and Cascade RCNN~\cite{cai2018cascade,cai2019cascade}, and the one-stage framework RetinaNet~\cite{retinaNet}. We follow the common practice in mmdetection~\cite{mmdetection}, \ie, multi-scale training with an AdamW optimizer and a batch size of 16. The initial learning rate is 0.0001 and the weight decay is 0.05. We adopt both 1$\times$ (12 epochs) and 3$\times$ (36 epochs) training schedules for the Mask RCNN framework to evaluate the object detection performance w.r.t. different backbones. For RetinaNet and Cascade RCNN, the models are trained with 1$\times$ and 3$\times$ schedules, respectively. The results on other settings are reported in the supplementary.

\begin{table}[htbp]
  \centering
  \caption{Object detection results on MS COCO with Mask RCNN.}
  \scriptsize
   \begin{threeparttable}
    \setlength{\tabcolsep}{0.001\linewidth}{\begin{tabular}{l|c|ccc|ccc|ccc|ccc}
    \hline
          & {Params} & \multicolumn{6}{c|}{Mask RCNN 1x}             & \multicolumn{6}{c}{Mask RCNN 3x} \\
          & (M) & AP$^{bb}$ & AP$_{50}^{bb}$ & AP$_{75}^{bb}$ & AP$^{mk}$ & AP$_{50}^{mk}$ & AP$_{75}^{mk}$ & AP$^{bb}$ & AP$_{50}^{bb}$ & AP$_{75}^{bb}$ & AP$^{mk}$ & AP$_{50}^{mk}$ & AP$_{75}^{mk}$ \\
    \hline
    ResNet50~\cite{he2016deep} & 44  & 38.6  & 59.5  & 42.1  & 35.2  & 56.3  & 37.5  & 40.8 &  61.2 &  44.4 & 37.0 &    58.4 & 39.3 \\
    ViL-S~\cite{yu2016multi} & 45  & 44.9  & 67.1  & 49.3  & 41.0  & 64.2  & 44.1  & 47.1  & 68.7  & 51.5  & 42.7  & 65.9  & 46.2  \\
    PVT-M~\cite{wang2021pyramid} & 64  & 42.0  & 64.4  & 45.6  & 39.0  & 61.6  & 42.1  & -     & -     & -     & -     & -     & - \\
    PVT-L~\cite{wang2021pyramid} & 81  & 42.9  & 65.0  & 46.6  & 39.5  & 61.9  & 42.5  &       &       &       &       &       &  \\
    PVTv2-B2~\cite{wang2021pvtv2} & 45  & 45.3  & 67.1  & 49.6  & 41.2  & 64.2  & 44.4  & -     & -     & -     & -     & -     & - \\
    CMT-S~\cite{guo2021cmt} & 45  & 44.6  & 66.8  & 48.9  & 40.7  & 63.9  & 43.4  & -     & -     & -     & -     & -     & - \\
    RegionViT-S~\cite{chen2021regionvit} & 50  & 42.5  & -     & -     & 39.5  & -     & -     & 46.3  & -     & -     & 42.3  & -     & - \\
    XCiT-S12/16~\cite{el2021xcit} & 44  & -     & -     & -     & -     & -     & -     & 45.3  & 67.0  & 49.5  & 40.8  & 64.0  & 43.8  \\
    DPT-M~\cite{chen2021dpt} & 66  & 43.8  & 66.2  & 48.3  & 40.3  & 63.1  & 43.4  & 44.3  & 65.6  & 48.8  & 40.7  & 63.1  & 44.1  \\
    ResT-Base~\cite{zhang2021rest} & 50  & 41.6  & 64.9  & 45.1  & 38.7  & 61.6  & 41.4  & -     & -     & -     & -     & -     & - \\
    Shuffle-T~\cite{huang2021shuffle} & 48 & - & - & - & - & - & - & 46.8 & 68.9 & 51.5 & 42.3 & 66.0 & 45.6 \\
    Focal-T~\cite{yang2021focal} & 49 & 44.8 & - & - & 41.0 & - & - & 47.2 & 69.4 & 51.9 & 42.7 & 66.5 & 45.9 \\
    \hline
    Swin-T~\cite{liu2021swin} & 48  & 43.7  & 66.6 & 47.7 & 39.8  & 63.3 & 42.7 & 46.0  & 68.1  & 50.3 & 41.6  & 65.1  & 44.9 \\
    \textbf{Swin-T+VSA} & \textbf{48}  & \textbf{45.6}  & \textbf{68.4}  & \textbf{50.1} & \textbf{41.4}  & \textbf{65.2} & \textbf{44.4} & \textbf{47.5}  & \textbf{69.4} & \textbf{52.3} & \textbf{42.8}  & \textbf{66.3} & \textbf{46.0} \\
    \hline
    ViTAEv2-S\footnotemark[1]~\cite{zhang2022vitaev2} & 39 & 43.5  & 65.8 & 47.4 & 39.4  &  62.6 & 41.8 &  44.7  & 65.8   & 49.1  & 40.0  &  62.6 & 42.8  \\
    \textbf{ViTAEv2-S\footnotemark[1]+VSA} & \textbf{39} & \textbf{45.9}  & \textbf{68.2} & \textbf{50.4} & \textbf{41.4}  & \textbf{65.1} & \textbf{44.5}  &  \textbf{48.1}  & \textbf{69.8}  & \textbf{52.9}  &  \textbf{42.9} & \textbf{66.9}  & \textbf{46.2} \\
    \hline
    \end{tabular}}%
  \label{tab:MaskRCNN}%
    \begin{tablenotes}
    \scriptsize
    \item[1] The full window version.
  \end{tablenotes}
  \end{threeparttable}
\end{table}%

\noindent\textbf{Results.} The results of baseline models and those with VSA on the MS COCO dataset with Mask RCNN, RetinaNet, and Cascade RCNN are reported in Tables~\ref{tab:MaskRCNN}, \ref{tab:RetinaNet}, and \ref{tab:Cascade}, respectively. Compared to the baseline method Swin-T~\cite{liu2021swin} and ViTAEv2~\cite{zhang2022vitaev2}, their VSA variants obtain better performance on both object detection and instance segmentation tasks with all detection frameworks, \eg, VSA brings a gain of 1.9 and 2.4 mAP$^{bb}$ for Swin-T and ViTAEv2-S with Mask RCNN 1$\times$ training schedule, confirming that VSA learns better object features than the vanilla window attention via the varied-size window attention that can better deal with objects at different scales for object detection. Besides, a longer training schedule (3$\times$) also sees a significant performance gain from VSA over the vanilla window attention. For example, the performance gain of VSA on Swin-T and ViTAEv2 reaches 1.5 mAP$^{bb}$ and 3.4 mAP$^{bb}$, respectively. We attribute this to the better attention regions learned by the VSR module in our VSA under longer training epochs. Similar conclusions can also be drawn when using RetinaNet~\cite{retinaNet} and Cascade RCNN~\cite{cai2018cascade} as detection frameworks, where VSA brings a gain of at least 2.0 and 1.2 mAP$^{bb}$, respectively. It is also noteworthy that the performance gains on ViTAEv2 are more significant than those on Swin-T. This is because there is no shifted window mechanism existing in ViTAEv2, and thus the ability to model long-range dependencies via attention is constrained within each window. In contrast, the varied size window attention in VSA empowers ViTAEv2 models to have such an ability and efficiently exchange rich contextual information across windows. 

\thisfloatsetup{heightadjust=all,valign=c}
\begin{table}[htbp]
  \centering
  \begin{floatrow}[2]
  \tablebox{
  \caption{Object detection results on MS COCO~\cite{lin2014microsoft} with RetinaNet~\cite{retinaNet}.}\label{tab:RetinaNet}%
  }{
  \scriptsize
    \setlength{\tabcolsep}{0.001\linewidth}{
    \begin{threeparttable}
    \begin{tabular}{l|c|ccc}
    \hline
        & Params & \multicolumn{3}{c}{RetinaNet} \\
        \cline{3-5}
          & (M) & AP$^{bb}$ & AP$_{50}^{bb}$ & AP$_{75}^{bb}$ \\
    \hline
    ResNet50~\cite{he2016deep} & 38 & 36.3 & 55.3 & 38.6 \\
    PVTv2-B1~\cite{wang2021pvtv2} & 24 & 41.2 & 61.9 & 43.9 \\
    ResT-Base~\cite{zhang2021rest} & 41 & 42.0 & 63.2 & 44.8 \\
    DAT-T~\cite{xia2022vision} & 39 & 42.8 & 64.4 & 45.2 \\
    Twins-SVT-S~\cite{chu2021twins} & 34 & 42.3 & 63.4 & 45.2 \\
    \hline
    Swin-T~\cite{liu2021swin} & 39 & 41.6 & 62.1 & 44.2 \\
    \textbf{Swin-T+VSA} & \textbf{39} & \textbf{43.6} & \textbf{64.8}  &  \textbf{46.6} \\
    \hline
    ViTAEv2-S\footnotemark[1]~\cite{zhang2022vitaev2} & 30 &  42.1 &  62.7 & 44.8 \\
    \textbf{ViTAEv2-S\footnotemark[1]+VSA} & \textbf{30} &  \textbf{44.3} &  \textbf{65.2}  & \textbf{47.6} \\
    \hline
    \end{tabular}
    \begin{tablenotes}
    \scriptsize
    \item[1] The full window version.
  \end{tablenotes}
  \end{threeparttable}
    }}%
    \tablebox{
    \caption{Object detection results on MS COCO~\cite{lin2014microsoft} with Cascade RCNN~\cite{cai2018cascade}.}\label{tab:Cascade}
    }{
    \scriptsize
    \begin{threeparttable}
    \begin{tabular}{l|c|ccc}
    \hline
        & Params & \multicolumn{3}{c}{Cascade RCNN} \\
    \cline{3-5}
          & (M)& AP$^{bb}$ & AP$_{50}^{bb}$ & AP$_{75}^{bb}$ \\
    \hline
    ResNet50~\cite{he2016deep} & 82  & 44.3  & 62.4  & 48.5  \\
    PVTv2-B2~\cite{wang2021pvtv2}] & 83  & 51.1  & 69.8  & 55.3  \\
    PVTv2-B2-Li~\cite{wang2021pvtv2} & 80  & 50.9  & 69.5  & 55.2  \\
    MSG-T~\cite{fang2021msg} & 83  & 51.4  & 70.1  & 56.0  \\
    \hline
    Swin-T~\cite{liu2021swin} & 86  & 50.2  & 68.8  & 54.7 \\
    \textbf{Swin-T+VSA} & \textbf{86}  & \textbf{51.4}  & \textbf{70.4}  & \textbf{55.9} \\
    \hline
    ViTAEv2-S\footnotemark[1]~\cite{zhang2022vitaev2} & 77  & 48.0  & 65.7  & 52.5 \\
    \textbf{ViTAEv2-S\footnotemark[1]+VSA} & \textbf{77}  & \textbf{51.9}  & \textbf{70.6}  & \textbf{56.2} \\
    \hline
    \end{tabular}
    \begin{tablenotes}
    \scriptsize
    \item[1] The full window version.
  \end{tablenotes}
  \end{threeparttable}
    }%
    \end{floatrow}
    
\end{table}%

\subsection{Semantic segmentation on Cityscapes}

\textbf{Settings.} The Cityscapes~\cite{Cordts2016Cityscapes} dataset is adopted to evaluate the performance of different backbones for semantic segmentation. The dataset contains over 5K well-annotated images of street scenes from 50 different cities. UperNet~\cite{xiao2018unified} is adopted as the segmentation framework. The training and evaluation of the models follow the common practice, \ie, using the Adam optimizer with polynomial learning rate schedulers. The models are trained for 40k iterations and 80k iterations separately with both 512$\times$1024 and 769$\times$769 input resolutions.

\begin{table}[htbp]
  \centering
  \scriptsize
  \caption{Semantic segmentation results on Cityscapes~\cite{Cordts2016Cityscapes} with 
  UperNet~\cite{xiao2018unified}. * denotes results are obtained with multi-scale test.}
    \setlength{\tabcolsep}{0.0028\linewidth}\begin{tabular}{l|ccc|ccc|ccc|ccc}
    \hline
          & \multicolumn{6}{c|}{512$\times$1024}                  & \multicolumn{6}{c}{769$\times$769} \\
\cline{2-13}          & \multicolumn{3}{c|}{40k} & \multicolumn{3}{c|}{80k} & \multicolumn{3}{c|}{40k} & \multicolumn{3}{c}{80k} \\
\cline{2-13}          & mIoU  & mAcc  & mIoU* & mIoU  & mAcc  & mIoU* & mIoU  & mAcc  & mIoU* & mIoU  & mAcc  & mIoU* \\
    \hline
    ResNet50~\cite{he2016deep} & 77.1  & 84.0  & 78.4     & 78.2  & 84.6  & 79.2 & 78.0  & 86.7  & 79.7     & 79.4  & 87.2  & 80.9 \\
    \hline
    Swin-T~\cite{liu2021swin} & 78.9  & 85.3  & 79.9  & 79.3  & 85.7  & 80.2  & 79.3  & 86.7  & 79.8  & 79.6  & 86.6  & 80.1  \\
        \textbf{Swin-T+VSA} & \textbf{80.8}  & \textbf{87.6}  & \textbf{81.7}  & \textbf{81.5}  & \textbf{87.8}  & \textbf{82.4}  & \textbf{81.0}  & \textbf{88.0}  & \textbf{81.9}  & \textbf{81.6}  & \textbf{88.3}  & \textbf{82.5}  \\
    \hline
    ViTAEv2-S~\cite{zhang2022vitaev2} & 80.1 & 86.5 & 80.9 & 80.8 & 87.0 & 81.0 & 79.6 & 86.1 & 80.6 & 80.5 & 86.8 & 81.2 \\
    \textbf{ViTAEv2-S+VSA} & \textbf{81.4} & \textbf{87.9} & \textbf{82.3} & \textbf{82.2} & \textbf{88.6} & \textbf{83.0} & \textbf{80.6} & \textbf{87.1} & \textbf{81.4} & \textbf{81.5} & \textbf{88.0} & \textbf{82.4} \\
    \hline
    Swin-S~\cite{liu2021swin} & 80.7 & 87.3 & 82.0 & 81.2 & 87.4 & 82.2 & 80.9 & 87.8 & 81.6 & 81.5 & 88.1 & 82.3 \\
    \textbf{Swin-S+VSA} & \textbf{82.1} & \textbf{88.5} & \textbf{83.2} & \textbf{82.8} & \textbf{88.9} & \textbf{83.6} & \textbf{82.0} & \textbf{88.7} & \textbf{83.0} & \textbf{82.8} & \textbf{89.5} & \textbf{83.6} \\
    \hline
    \end{tabular}%
  \label{tab:cityscape}%
\end{table}%
\noindent\textbf{Results.} The results are available in Table~\ref{tab:cityscape}. With 512$\times$1024 input size, VSA brings over 1.3 mIoU and 1.4 mAcc gains for both Swin-T~\cite{liu2021swin} and ViTAEv2-S~\cite{zhang2022vitaev2}, no matter with 40k or 80k training schedules. This observations hold with 769$\times$769 resolution images as input, where VSA brings over 1.0 mIoU and 1.0 mAcc gains for both models. Such phenomena validates the effectiveness of the proposed VSA in improving the baseline models' performance on semantic segmentation tasks. With more training iterations (80k), the performance gains of VSA over Swin-T increases from 1.9 to 2.2 mIoU with 512$\times$1024 and from 1.7 to 2.0 mIoU with 769$\times$769, owing to the better attention regions learned by the VSR module.
Besides, with multi-scale testing, the performance of using VSA further improves, indicating that VSA can implicit capture multi-scale features as the target windows have different scales and locations for each head.

\subsection{Ablation Study}

We adopt Swin-T~\cite{liu2021swin} with VSA for ablation studies. The models are trained for 300 epochs with AdamW optimizer. To find the optimal configuration of VSA, we gradually substitute the window attention in different stages of Swin with VSA. The results are shown in Table~\ref{tab:ablation1}, where $\checkmark$ indicates that VSA replaces the vanilla window attention. We can see that the performance gradually improves with more VSA used and reaches the best when using VSA in all four stages. Meanwhile, it only takes a few extra parameters and FLOPs. Therefore, we choose to use VSA at all stages as the default setting in this paper.

\begin{table}[htbp]
  \centering
  \begin{floatrow}[2]
  \tablebox{
  \caption{The ablation study of using VSA in each stage of Swin-T~\cite{liu2021swin}.}\label{tab:ablation1}
  }{
  \scriptsize
  \begin{tabular}{cccc|cc|c}
    \hline
    \multicolumn{4}{c|}{VSA at stages} & FLOPs & Param & Acc. \\
\cline{1-4}    stage 1 & stage 2 & stage 3 & stage 4 & (G)   & (M)   & (\%) \\
    \hline
          &       &       &       & 4.5   & 28.2  & 81.2 \\
          &       &       & \checkmark & 4.5   & 28.3  & 81.4 \\
          &       & \checkmark & \checkmark & 4.6   & 28.7  & 81.9 \\
          & \checkmark & \checkmark & \checkmark & 4.6   & 28.7  & 82.1 \\
    \checkmark & \checkmark & \checkmark & \checkmark & 4.6   & 28.7  & 82.3 \\
    \hline
    \end{tabular}%
  }
  \tablebox{
  \caption{The ablation study of each component in VSA based on Swin-T~\cite{liu2021swin}.}\label{tab:ablation2}
  }{
  \scriptsize
  \begin{tabular}{ccc|c}
    \hline
    CPE   & VSR & Shift & Acc. \\
    \hline
          &       & \checkmark     & 81.2 \\
          & \checkmark     &       & 81.6 \\
    \checkmark     &       & \checkmark     & 81.6 \\
    \checkmark     & \checkmark     &       & 82.3 \\
    \checkmark     & \checkmark     & \checkmark     & 82.3 \\
    \hline
    \end{tabular}%
  }
  \end{floatrow}
\end{table}%

We take Swin-T as the baseline and further validate the contribution of each component in VSA. The results are available in Table~\ref{tab:ablation2}, where $\checkmark$ denotes using the specific component. `Shift' is short for the shifted window mechanism. With only `Shift' marked, the model becomes the baseline Swin-T. As can be seen, the model with `VSR' alone outperforms Swin-T by 0.3\% absolute accuracy, implying (1) the effectiveness of varied-size windows in cross-window information exchange and (2) the advantage of adapting the window sizes and locations, \ie, attention regions, to the objects at different scales. Besides, using CPE and VSR in VSA further boosts to 82.3\%, which outperforms the variant of `CPE' + `Shift' by 0.6\% accuracy. It indicates that CPE is better compatible with varied size windows by providing local positional information. It is also noteworthy that there is no need to use the shifted-window mechanism in VSA according to the results in the last two rows, confirming that varied-size windows can guarantee the feature exchange across overlapped windows.

\subsection{Throughputs \& GPU memory comparison}
\begin{table}[htbp]
  \centering
  \scriptsize
  \caption{Throughput \& GPU memory comparison with VSA.}
    \begin{tabular}{c|c|c|c}
    \hline
          & Throughputs on A100 & Throughputs on V100 & Memory  \\
          & (fps) & (fps) & (G) \\
    \hline
    Swin-T & 1557 & 679 & 15.8 \\
    Swin-T+VSA & 1297 & 595 & 16.1 \\
    \hline
    Swin-S & 961 & 401 & 23.0 \\
    Swin-S+VSA & 769 & 352  & 23.5 \\
    \hline
    \end{tabular}%
  \label{tab:Speed}%
\end{table}%

We also evaluate the model's throughputs during inference and GPU memory consumption during training, with batch size 128 and input resolution 224 $\times$ 224. We run each model 20 times firstly as warmup and count the average throughputs of the subsequent 30 runs as the throughputs of the models. All of the experiments are conducted on the NVIDIA A100 and V100 GPUs. As shown in Table~\ref{tab:Speed}, VSA slows down the Swin model by about 12\%$\sim$17\% on different hardware platforms and consumes 2\% more GPU memory, with much better performance on both classification and downstream dense prediction tasks. Such slow-down and extra memory consumption is mainly due to the sub-optimal optimization of sampling operations compared with the matrix multiply operations in the PyTorch framework, where the latter is sufficiently optimized with cuBLAS. Integrating the sampling operation with following linear projection operations with CUDA optimization can help alleviate the speed concerns, which we leave as our future work to implement the proposed VSR module better. 

\subsection{Visual inspection and analysis}

\begin{figure}
    \centering
    \includegraphics[width=\linewidth]{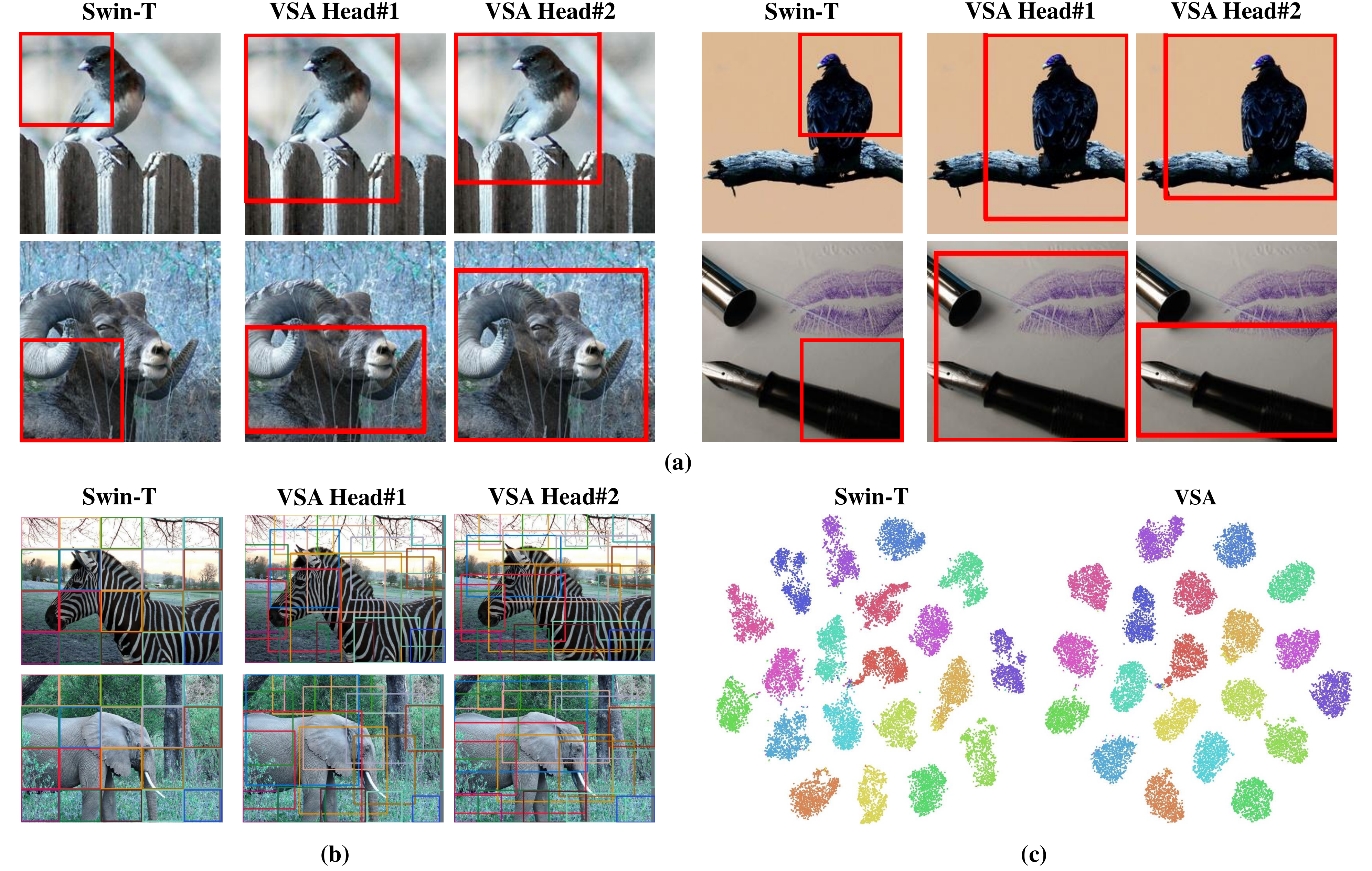}
    \caption{Visualization of the varied-size windows generated by VSA from ImageNet (a) and MS COCO (b). The t-SNE analysis is also provided in (c).}
    \label{fig:variedWin}
\end{figure}

\noindent\textbf{Visualization of target windows.} We visualize the default windows used in Swin-T~\cite{liu2021swin} and the varied-size windows generated by VSA on images from the ImageNet~\cite{deng2009imagenet} and MS COCO~\cite{lin2014microsoft} datasets to see where VSA learns to attend for different images. The results are visualized in Figure~\ref{fig:variedWin}. As shown in Figure~\ref{fig:variedWin}(a), the generated windows from VSA can better cover the target objects in the images while the fixed-size windows adopted in Swin can only capture part of the targets. It can also be inferred from Figure~\ref{fig:variedWin}(b) that the windows generated by different heads in VSA have different sizes and locations to focus on different parts of the targets, which helps to capture rich contextual information and learn better object feature representations. Besides, the windows that cover the target objects have more variance in size and location compared with those covering background as shown in (b), \eg, the windows on the zebra and elephant vary (the blue, red, orange, pink, \etc) significantly while others in the background are less varied. In addition, the target windows are overlapped with each other, thus enabling abundant cross-window feature exchange and making it possible to drop the shifted window mechanism in VSA.

\noindent\textbf{t-SNE analysis.} We further use t-SNE to analyze the features generated by Swin-T models with and without VSA. We randomly select 20 categories from the ImageNet dataset and use t-SNE to visualize the extracted features. As shown in Figure~\ref{fig:variedWin}(c), the features generated by Swin-T with VSA are better clustered, demonstrating that VSA can help the models deal with objects of different sizes and learn more discriminative features.

\section{Limitation and Discussion}

Although VSA has been proven efficient in dealing with images of varied resolutions and has shown its effectiveness on various vision tasks, including classification, detection, instance segmentation, and semantic segmentation, we only evaluate VSA with Swin~\cite{liu2021swin} and ViTAEv2~\cite{zhang2022vitaev2} in this paper. It will be our future work to explore the usage of VSA on other transformers with window-based attentions, \eg, CSwin~\cite{dong2021cswin} and Pale~\cite{wu2021pale}, which use cross-shaped attentions. Besides, to keep the computational cost as the vanilla window attention, we only sample sparse tokens from each target window, \ie, the number of sampled tokens equals the default window size, which may ignore some details when the window becomes extremely large. Although the missed details may be complemented from other windows via feature exchange, a more efficient sampling strategy can be explored in the future study.

\section{Conclusion}
This paper presents a novel varied-size window attention (VSA), \ie, an easy-to-implement module that can help boost the performance of representative window-based vision transformers such as Swin in various vision tasks, including image classification, object detection, instance segmentation, and semantic segmentation. By estimating the appropriate window size and location for each image in a data-driven manner, VSA enables the transformers to attend to far-away yet relevant tokens with negligible extra computational cost, thereby modeling long-term dependencies among tokens, capturing rich context from diverse windows, and promoting information exchange among overlapped window. In the future, we will investigate the usage of VSA in more attentions types including cross-shaped windows, axial attentions, and others as long as they can be parameterized w.r.t. size (\eg, height, width, or radius), rotation angle, and position. We hope that this study can provide useful insight to the community in developing more advanced attention mechanisms as well as vision transformers.

\noindent \textbf{Acknowledgement} Mr. Qiming Zhang, Mr. Yufei Xu, and Dr. Jing Zhang are supported by ARC FL-170100117.

\clearpage
%
%
\bibliographystyle{splncs04}
\bibliography{egbib}
\end{document}


\pagestyle{headings}
\mainmatter
\def\ECCVSubNumber{6686}  

\title{VSA: Learning Varied-Size Window Attention in Vision Transformers\\
Supplementary Material} 

\titlerunning{VSA}
%
\author{
Qiming Zhang\inst{1}*\orcidlink{0000-0003-0060-0543} \thanks{Equal contribution.} \and
Yufei Xu\inst{1}*\orcidlink{0000-0002-9931-5138} \and
Jing Zhang\inst{1}\orcidlink{0000-0001-6595-7661} \and 
Dacheng Tao\inst{2,1}\orcidlink{0000-0001-7225-5449}}

\authorrunning{Q. Zhang et al.}
%
\institute{University of Sydney, Australia \and
JD Explore Academy, China \\
\email{\{yuxu7116,qzha2506\}@uni.sydney.edu.au, \\ jing.zhang1@sydney.edu.au, dacheng.tao@gmail.com}}

\maketitle

\appendix

\section{Appendix}

\begin{figure}
    \centering
    \includegraphics[width=\linewidth]{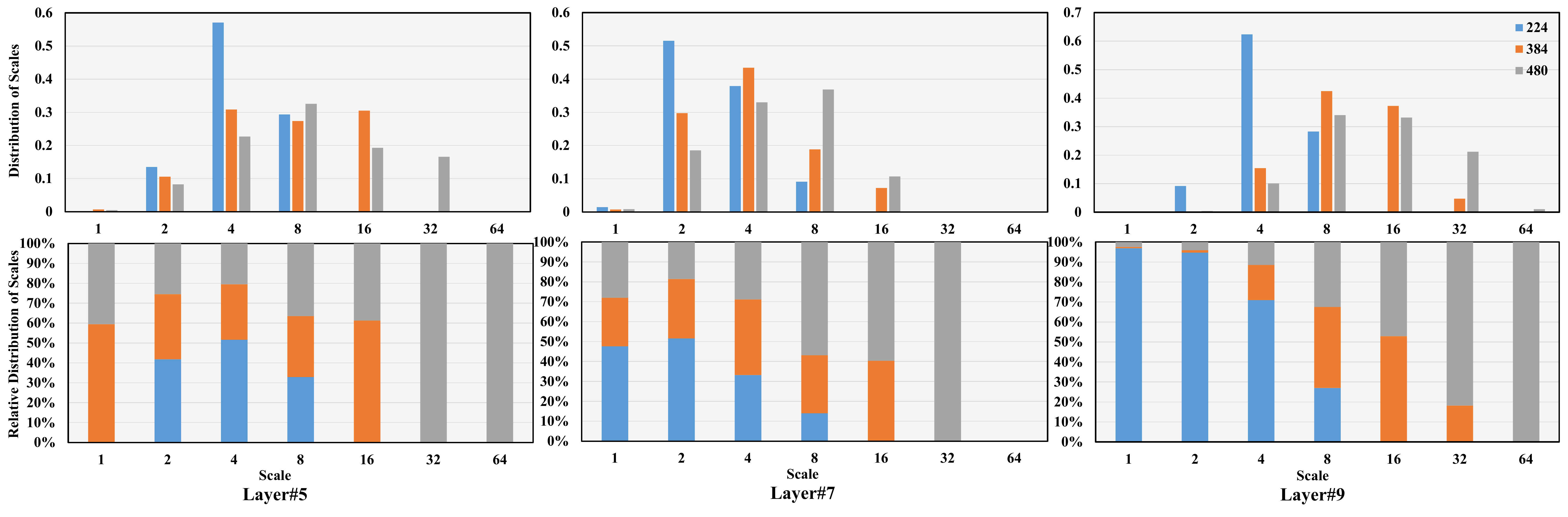}
    \caption{The absolute and relative distribution of scales estimated by VSA with different input resolutions. The X-axis denotes the scale ratio of the varied-size window w.r.t. the default one. VSA generates the target windows at various scales to capture rich contextual information and the window size tends to become larger to adapt to large objects with the resolution going higher.}
    \label{fig:scaleDis}
\end{figure}

\subsection{Implementation details}
In this section we give details of the regression and sampling process. Different from previous work~\cite{chen2021dpt} that controls the regression distance by a predefined parameter, the proposed VSR module is hyper-parameter free. Denoting the coordinates of samples within one window as $\{(x_i,y_i) | i=1\dots n\}$ where $i$ refers to the $i_{th}$ token, we first disentangle the coordinates as
\begin{equation}
    \begin{split}
        x_i &= x^{re}_i + x^{ce}, \\
        y_i &= y^{re}_i + y^{ce},
    \end{split}
\end{equation}
where $(x^{ce},y^{ce})$ denote the center coordinates of the corresponding window and $(x^{re}_i,y^{re}_i)$ are the relative coordinates w.r.t. the center. Then given the learned scales and offsets $s_w,o_w \in R^{2}$ of the corresponding window from Equation 3, they are normalized by multiplying the ratio between the window size and image size (also denoted as $s_w,o_w$ for simplicity). This makes the VSR module learn how to expand and move the current window towards the optimal attention region by taking the window as a base. Then the tokens' new coordinates within that window are calculated as
\begin{equation}
    \begin{split}
        (x_i^*, y_i^*)^T = (x^{re}_i, y^{re}_i)^T \cdot s_w + o_w + (x^{ce}, y^{ce})^T.
    \end{split}
\end{equation}
In the end, the key and value tokens are sampled according to the new coordinates $(x_i^*, y_i^*)$ and fed into the following window-based attention layer. 

\begin{table}[ht]
  \centering
  \caption{Image classification results on ImageNet. `Input Size' denotes the image size used for training and test. `IN1k' and `IN22k' refer to ImageNet-1k and ImageNet-22k datasets respectively.}
  \scriptsize
  \begin{threeparttable}
    \begin{tabular}{l|cccc|cc|c}
    \hline
    \multicolumn{1}{c|}{\multirow{2}[2]{*}{Model}} & Params & FLOPs & Input & Training & \multicolumn{2}{c|}{ImageNet~\cite{deng2009imagenet}} & Real~\cite{beyer2020we} \\
          & (M)   & (G)   & Size & Set & Top-1 & Top-5 & Top-1 \\
    \hline
    DeiT-S~\cite{touvron2020training} & 22  & 4.6   & 224 & IN1k  & 81.2  & 95.4  & 86.8  \\
    PVT-S~\cite{wang2021pyramid} & 25  & 3.8   & 224 & IN1k   & 79.8  & - & -  \\
    ViL-S~\cite{zhang2021multi} & 25 & 4.9 & 224 & IN1k & 82.4 & - & - \\
    PiT-S~\cite{heo2021pit} & 24  & 4.8   & 224 & IN1k  & 80.9  &   -    & - \\
    TNT-S~\cite{han2021transformer} & 24  & 5.2   & 224  & IN1k  & 81.3  & 95.6  & - \\
    MSG-T~\cite{fang2021msg} & 25 & 3.8 & 224 & IN1k & 82.4 & - & - \\
    Twins-PCPVT-S~\cite{chu2021twins} & 24  & 3.8   & 224 & IN1k   & 81.2  &    -   &  - \\
    Twins-SVT-S~\cite{chu2021twins} & 24  & 2.9   & 224  & IN1k  & 81.7  & -      & - \\
    T2T-ViT-14~\cite{yuan2021tokens} & 22  & 5.2   & 224  & IN1k  & 81.5  & 95.7  & 86.8  \\
    Swin-T~\cite{liu2021swin} & 29  & 4.5   & 224 & IN1k   & 81.2  &    -   & - \\
    \textbf{Swin-T$+$VSA} & \textbf{29} & \textbf{4.6} & \textbf{224} & \textbf{IN1k} & \textbf{82.3} & \textbf{96.1}  & \textbf{87.5} \\
    ViTAEv2-S\footnotemark[1]~\cite{zhang2022vitaev2} &  20  & 5.4  & 224  & IN1k & 82.2  & 96.1  & 87.5 \\
    ViTAEv2-S~\cite{zhang2022vitaev2} & 20 & 5.7 & 224 & IN1k & 82.6 & 96.2 & 87.6 \\
    \textbf{ViTAEv2-S\footnotemark[1]$+$VSA} &  \textbf{20}   &  \textbf{5.6} & \textbf{224} & \textbf{IN1k} & \textbf{82.7} &  \textbf{96.3} & \textbf{87.8} \\
    \hline
    Swin-T~\cite{liu2021swin} & 29  &  14.2  & 384  & IN1k  & 81.4  &  95.4 & 86.4 \\
    \textbf{Swin-T$+$VSA} & \textbf{29} & \textbf{14.9} & \textbf{384} & \textbf{IN1k}  & \textbf{83.2} & \textbf{96.5}  & \textbf{88.0} \\
    Swin-T~\cite{liu2021swin} & 29  &  23.2  & 480  & IN1k & 81.5  &  95.7 & 86.3 \\
    \textbf{Swin-T$+$VSA} & \textbf{29} & \textbf{24.0} & \textbf{480}& \textbf{IN1k} & \textbf{83.4} & \textbf{96.7} & \textbf{88.0}  \\
    \hline
    PiT-B~\cite{heo2021pit} & 74  & 12.5  & 224 & IN1k   & 82.0  &   -    & - \\
    TNT-B~\cite{han2021transformer} & 66  & 14.1  & 224  & IN1k  & 82.8  & 96.3  & - \\
    Focal-B~\cite{yang2021focal} & 90  & 16.0  & 224 & IN1k   & 83.8  &  -     & - \\
    ViL-B~\cite{zhang2021multi} & 56 & 13.4 & 224 & IN1k & 83.7 & - & - \\
    MSG-S~\cite{fang2021msg} & 56 & 8.4 & 224 & IN1k & 83.4 & -  & - \\
    PVTv2-B5~\cite{wang2021pvtv2} & 82  & 11.8  & 224  & IN1k  & 83.8  &  -     & - \\
    Swin-S~\cite{liu2021swin} & 50  & 8.7   & 224  & IN1k  & 83.0  &   -    &  -\\
    Swin-B~\cite{liu2021swin} & 88  & 15.4  & 224  & IN1k  & 83.3  &  - & 88.0 \\
    Shuffle-S\cite{huang2021shuffle} & 50 & 8.9 & 224 & IN1k & 83.5 & - & - \\
    \textbf{Swin-S$+$VSA} & \textbf{50} & \textbf{8.9} & \textbf{224} & \textbf{IN1k}  & \textbf{83.6} &  \textbf{96.6} & \textbf{88.4} \\
    ViTAEv2-48M & 49 & 13.3 & 224 & IN1k & 83.8 & 96.6 & 88.4 \\
    \textbf{ViTAEv2-48M\footnotemark[1]$+$VSA} & \textbf{50} & \textbf{13.0} & \textbf{224} & \textbf{IN1k}  & \textbf{83.9} & - & - \\
    \hline
        Swin-B~\cite{liu2021swin} & 88  & 15.4  & 224   & 83.3  &  - & 88.0 \\
    \textbf{Swin-B$+$VSA} & 88 & 16.0 & 224 & 83.9 & 96.7 & 88.6 \\

    \hline
    \textbf{ViTAEv2-48M\footnotemark[1]$+$VSA} & \textbf{50} & \textbf{13.0} & \textbf{224} & \textbf{IN22k+IN1k} & \textbf{84.9} & - & - \\
    ViTAEv2-B & 90 & 24.3 & 224 & IN22k+IN1k & 86.1 & 97.9 & 89.9 \\
    \textbf{ViTAEv2-B\footnotemark[1]$+$VSA} & \textbf{94} & \textbf{23.9} & \textbf{224} & \textbf{IN22k+IN1k} & \textbf{86.2} & \textbf{97.9} & \textbf{90.0} \\
    \hline
    \end{tabular}%

  \begin{tablenotes}
    \scriptsize
    \item[1] The full window version.
  \end{tablenotes}
  \end{threeparttable}
  \label{stab:Classification}%
\end{table}%

\subsection{Visual inspection}

\noindent\textbf{Statistics of target windows' scales.} In this part, we analyze the behavior of VSA in regressing the target windows when dealing with input images of different resolutions. Specifically, we first scale the input images to different resolutions, \ie, 224$\times$224, 384$\times$384, and 480$\times$480 and train three corresponding models respectively, whose performance has been reported in Table~\ref{stab:Classification}. Then, we randomly sample images from the ImageNet validation set and feed them into the three models respectively to count the scale distribution of all target windows estimated by VSR. We plot the results from the 5th, 7th, and 9th layers in Figure~\ref{fig:scaleDis}, where the X-axis denotes the scale ratio between the target windows and the default one. The Y-axis in the first row denotes the percentage of corresponding scales. For better visualization, we normalize the data independently along the Y-axis w.r.t. each scale on the X-axis, \ie, obtaining the relative distribution as shown at the bottom of the figure. As can be seen, larger windows gradually dominate the VSA when the input size increases from 224$\times$224 to 480$\times$480, showing the good ability of VSA in adapting to different input sizes.

\subsection{Classification results}

We present the classification results of various model in Table~\ref{stab:Classification}. As shown in the table, the proposed VSA method can consistently improve the classification results. It is noted that with the help of VSA, the full window version of ViTAEv2 outperforms its counterpart using full attention in the last two stages~\cite{zhang2022vitaev2}, which has much computation cost when the input images become larger as shown in \cite{zhang2022vitaev2}. Besides, when pretraining using the ImageNet-22k dataset, the full window version of ViTAEv2-B reaches 86.2\% and 90.0\% Top-1 accuracy on ImageNet-1k and ImageNet-1k Real datasets respectively.

\begin{table}[htbp]
  \centering
  \caption{Object detection results on MS COCO with Cascade RCNN.}
  \scriptsize
  \begin{threeparttable}
    \setlength{\tabcolsep}{0.001\linewidth}{\begin{tabular}{l|c|ccc|ccc|ccc|ccc}
    \hline
          & Params & \multicolumn{6}{c|}{Cascade RCNN 1x}     & \multicolumn{6}{c}{Cascade RCNN 3x} \\
          & (M) & AP$^{bb}$ & AP$_{50}^{bb}$ & AP$_{75}^{bb}$ & AP$^{mk}$ & AP$_{50}^{mk}$ & AP$_{75}^{mk}$ & AP$^{bb}$ & AP$_{50}^{bb}$ & AP$_{75}^{bb}$ & AP$^{mk}$ & AP$_{50}^{mk}$ & AP$_{75}^{mk}$ \\
    \hline
    R50   & 82    & 44.3  & 62.7  & 48.4  & 38.3  & 59.7  & 41.2  & 46.3  & 64.3  & 50.5  & 40.1  & 61.7  & 43.4 \\
    Swin-T & 86    & 48.1  & 67.1  & 52.2  & 41.7  & 64.4  & 45.0  & 50.2  & 69.2  & 54.7  & 43.7  & 66.6  & 47.3 \\
    \textbf{Swin-T+VSA} & \textbf{86} & \textbf{49.8} & \textbf{69.0} & \textbf{54.1} & \textbf{43.0} & \textbf{66.2} & \textbf{46.4} & \textbf{51.3} & \textbf{70.3} & \textbf{55.8} & \textbf{44.6} & \textbf{67.6} & \textbf{48.1} \\
    \hline
    ViTAEv2-S\footnotemark[1] & 77    & 47.3  & 66.0  & 51.5  & 40.6  & 63.0  & 43.7  & 48.0  & 65.7  & 52.5  & 41.3  & 63.1  & 45.0 \\
    \textbf{ViTAEv2-S\footnotemark[1]+VSA} & \textbf{77} & \textbf{49.8} & \textbf{69.2} & \textbf{54.0} & \textbf{43.1} & \textbf{66.4} & \textbf{46.5} & \textbf{51.9} & \textbf{70.6} & \textbf{56.2} & \textbf{44.8} & \textbf{68.1} & \textbf{48.5} \\
    \hline
    \end{tabular}}%
  
  \begin{tablenotes}
    \scriptsize
    \item[1] The full window version.
  \end{tablenotes}
  \end{threeparttable}
  \label{stab:Cascade}%
\end{table}%

\begin{table}[htbp]
  \centering
  \scriptsize
  \setlength{\tabcolsep}{0.01\linewidth}{
    \begin{tabular}{c|c|ccc|c|ccc|c}
    \hline
          & \multicolumn{9}{c}{CascadeRCNN 3x} \\
    \cline{2-10}
          & $AP^{bb}$ & $AP^{bb}_{s}$ & $AP^{bb}_{m}$ & $AP^{bb}_{l}$ & $AP^{M}$ & $AP^{M}_{s}$ & $AP^{M}_{m}$ & $AP^{M}_{l}$ & fps\\
    \hline
    Swin-T & 50.4  & 33.8  & 54.1  & 65.2  & 43.7  & 27.3  & 47.5  & 59.0  & 25.2 \\
    Swin-T+VSA & 51.4  & 35.5  & 54.7  & 66.0  & 44.7  & 28.7  & 48.1  & 59.8  & 22.5 \\
    \hline
    Swin-S & 51.9  & 35.2  & 55.7  & 67.7  & 45.0  & 28.8  & 48.7  & 60.6  & 16.0 \\
    Swin-S+VSA & 52.7  & 36.7  & 56.0  & 68.3  & 45.6  & 29.8  & 49.0  & 61.2  & 14.1 \\
    \hline
    Swin-B & 51.9  & 35.4  & 55.2  & 67.4  & 45.0  & 28.9  & 48.3  & 60.4  & 14.4 \\
    Swin-B+VSA & 52.9  & 36.8  & 56.4  & 68.4  & 45.9  & 30.1  & 49.3  & 61.5  & 12.8 \\
    \hline
    \end{tabular}}
  \caption{More object detection results with Cascade RCNN.}
  \label{stab:More Cascade}
\end{table}%

\subsection{Downstream task results}

We further present the results of Swin~\cite{liu2021swin} and ViTAEv2~\cite{zhang2022vitaev2} with the proposed VSR module on MS COCO dataset~\cite{lin2014microsoft} for detection tasks. ViTAEv2 with window attention for all stages, \ie, the full window attention version, is adopted as default. We evaluate their detection performance Cascade RCNN~\cite{cai2018cascade} frameworks with both 1$\times$ and 3$\times$ settings. The detection results are available in Table~\ref{stab:Cascade} and Table~\ref{stab:More Cascade} (Cascade RCNN). The results in both tables imply that VSA can successfully boost the detection performance with Cascade RCNN frameworks. Besides, the performance gain keeps when scaling to models with more parameters, \eg, Swin-S and Swin-B~\cite{liu2021swin} with 50M and 88 parameters, respectively, as demonstrated in Table~\ref{stab:More Cascade}. In addition, from Table~\ref{stab:More Cascade}, we can see that the improvement on small and big objects significantly exceeds the median ones of the baseline, which supports our claim that VSA can better deal with objects of different sizes.

\clearpage
%
%
\bibliographystyle{splncs04}
\bibliography{egbib}